\documentclass[letterpaper]{article} 
\usepackage{aaai2026}  
\usepackage{times}  
\usepackage{helvet}  
\usepackage{courier}  
\usepackage[hyphens]{url}  
\usepackage{graphicx} 
\usepackage{natbib}  
\usepackage{caption} 
\usepackage{algorithm}
\usepackage{algorithmic}
\usepackage{booktabs}
\usepackage{colortbl}
\usepackage{multirow}
\usepackage{amssymb}
\usepackage{amsmath}
\usepackage{newfloat}
\usepackage{listings}
\DeclareCaptionStyle{ruled}{labelfont=normalfont,labelsep=colon,strut=off} 
\floatstyle{ruled}
\newfloat{listing}{tb}{lst}{}
\floatname{listing}{Listing}
\title{Complex Mathematical Expression Recognition: \\ Benchmark, Large-Scale Dataset and Strong Baseline}

\author{
    Weikang Bai\textsuperscript{\rm 1,}\equalcontrib,
    Yongkun Du\textsuperscript{\rm 1,}\equalcontrib,
    Yuchen Su\textsuperscript{\rm 1},
    Yazhen Xie\textsuperscript{\rm 1},
    Zhineng Chen\textsuperscript{\rm 2,}\thanks{Corresponding author.}
}

\affiliations{
\textsuperscript{\rm 1}College of Computer Science and Artificial Intelligence, Fudan University, Shanghai, China\\
\textsuperscript{\rm 2}Institute of Trustworthy Embodied AI, Fudan University, Shanghai, China\\
\{wkbai24, ykdu23, ycsu23, yzxie24\}@m.fudan.edu.cn, zhinchen@fudan.edu.cn
}

\usepackage{bibentry}

\begin{document}

\maketitle

\begin{abstract}
Mathematical Expression Recognition (MER) has made significant progress in recognizing simple expressions, but the robust recognition of complex mathematical expressions with many tokens and multiple lines remains a formidable challenge. In this paper, we first introduce CMER-Bench, a carefully constructed benchmark that categorizes expressions into three difficulty levels: easy, moderate, and complex. Leveraging CMER-Bench, we conduct a comprehensive evaluation of existing MER models and general-purpose multimodal large language models (MLLMs). The results reveal that while current methods perform well on easy and moderate expressions, their performance degrades significantly when handling complex mathematical expressions, mainly because existing public training datasets are primarily composed of simple samples. In response, we propose MER-17M and CMER-3M that are large-scale datasets emphasizing the recognition of complex mathematical expressions. The datasets provide rich and diverse samples to support the development of accurate and robust complex MER models. Furthermore, to address the challenges posed by the complicated spatial layout of complex expressions, we introduce a novel expression tokenizer, and a new representation called Structured Mathematical Language, which explicitly models the hierarchical and spatial structure of expressions beyond LaTeX format. Based on these, we propose a specialized model named CMERNet, built upon an encoder-decoder architecture and trained on CMER-3M. Experimental results show that CMERNet, with only 125 million parameters, significantly outperforms existing MER models and MLLMs on CMER-Bench.
\end{abstract}



\begin{links}
    \link{Code}{https://github.com/Baitlo/CMER}
\end{links}

\begin{figure}[t]
\centering 
    \includegraphics[width=0.48\textwidth]{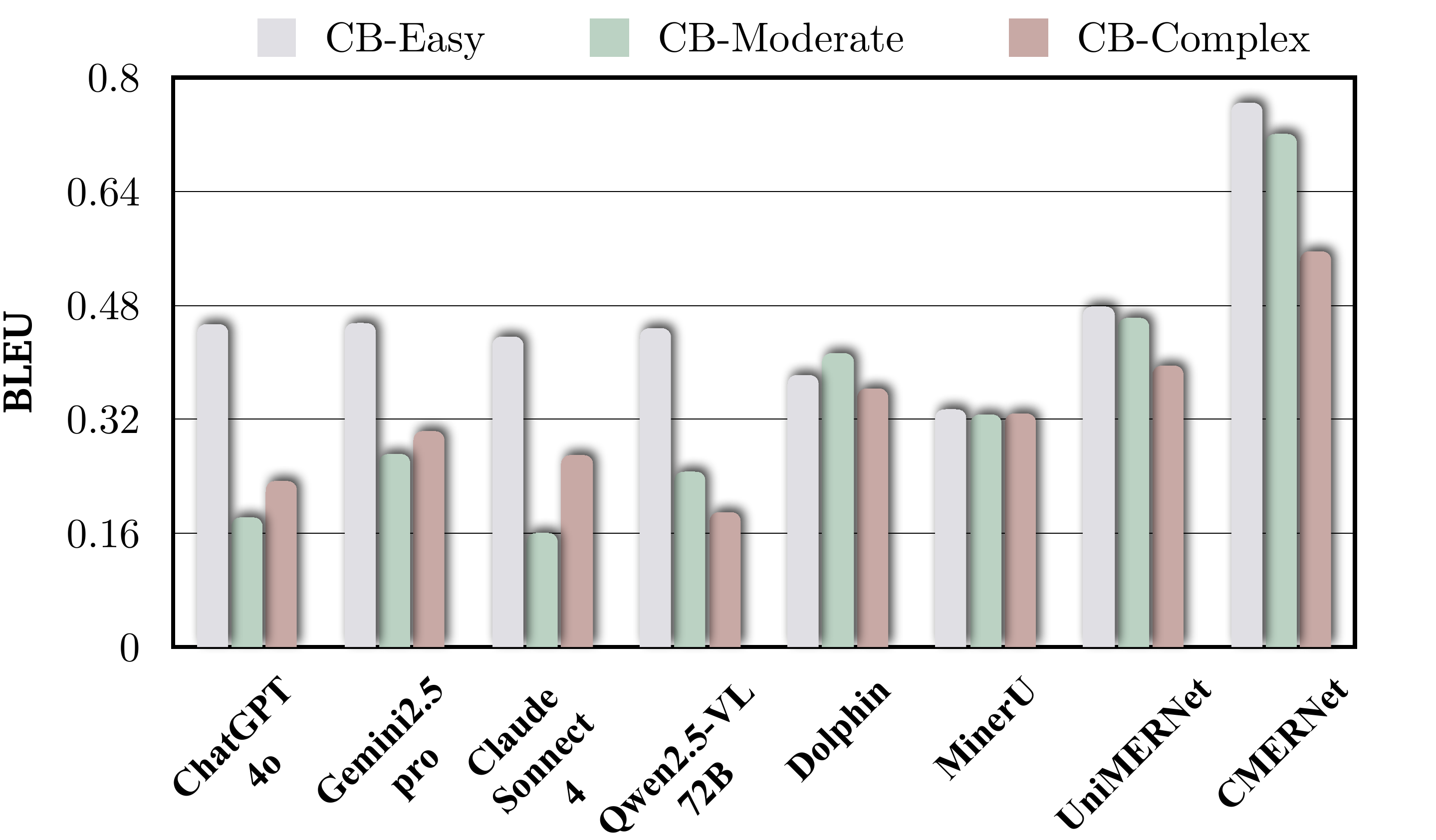}
\caption{Comparing our CMERNet with existing models in complex mathematical expression recognition. CB means CMER-Bench, our constructed benchmark.}
\label{fig:1}
\end{figure}
\section{introduction}

Mathematical Expression Recognition (MER) has long been a fundamental task for visual-based document parsing and understanding. Unlike Optical Character Recognition (OCR) that primarily deals with plain text~\cite{du2022svtr,zheng2024cdistnet,du2025context,du2025instruction,du2025svtrv2}, MER poses unique challenges due to its two-dimensional spatial layout, diverse symbols, complex syntax, and high structural ambiguity. In recent years, a variety of specialized methods have emerged to tackle this problem, such as \cite{deng2017im2markup,yuan2022syntax,zhu_tamer_2024,wang_unimernet_2024,li2025pacm}, which have significantly advanced MER.

Existing approaches can be broadly classified into two paradigms:
the first is the image-to-sequence interpretation, which treats MER as a direct image-to-LaTeX translation task. One representative work in this category is UniMER~\cite{wang_unimernet_2024}, which leverages a custom visual encoder and an LLM decoder to autoregressively generate a LaTeX string from the input image. This work also proposes a dataset over one million scientific expressions for model training~\cite{wang_unimernet_2024}. The second paradigm focuses on explicitly modeling the 2D structure of expressions, typically applied to handwritten MER. A prominent example is TDv2 by \citeauthor{Wu_tdv2_2022}. Instead of generating a linear LaTeX string, TDv2 employs a tree-structured decoder that autoregressively constructs a syntax tree for the expression. At each step, it decouples the prediction into two sub-tasks: a Node Classification Module to identify the next symbol, and a Branch Prediction Module to determine its spatial relationship (e.g., superscript, below) to its parent node. In this manner, domain knowledge of expressions is incorporated for recognition.


\begin{figure*}[!t]
\centering 
    \includegraphics[width=1.0\textwidth]{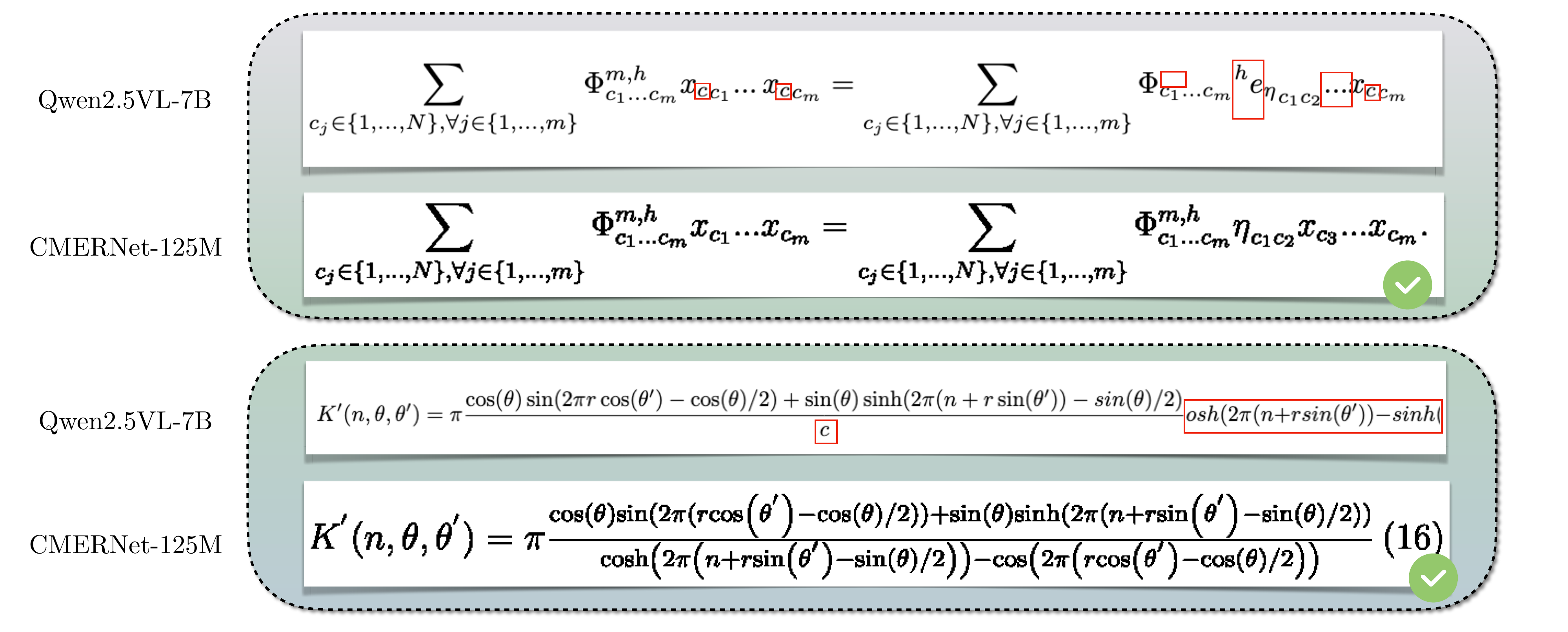}
\caption{Visualization comparison of MER results between Qwen2.5VL and our CMERNet.}
\label{fig:3}
\end{figure*}


Additionally, recent advances built on Multimodal Large Language Models (MLLMs) have also made remarkable progress in MER~\cite{wang_unimernet_2024,Blecher2022latexocr}.
However, due to the simple nature of current MER benchmarks~\cite{deng2017im2markup,wang_unimernet_2024}, both specialized methods and general-purpose MLLMs are biased towards mathematical expressions of limited complexity, typically involving dozens of tokens and a single-line layout. As shown in Fig.~\ref{fig:3}, these methods falter when confronted with complex, multi-line expressions commonly seen in scientific documents. Motivated by this observation, we argue that current MER models face great challenges in complex scenarios, which can be summarized as the following three fundamental challenges:

\textbf{Lack of Challenging Benchmarks.} Existing benchmarks predominantly focus on evaluating simple expressions. There is a lack of complex expressions containing hundreds of tokens, multiple lines, intricate spatial layout, diverse symbols, etc. These features are essential for a comprehensive assessment of MER models. 

\textbf{Training Data Scarcity and Skewness.} MER currently suffers from a scarcity of large-scale training data. Existing datasets are not only limited in size but are also heavily skewed towards simple expressions. The distribution of expression length is highly imbalanced, with a critical underrepresentation of long and complex examples, which are critical to train robust MER models. 

\textbf{Data Representation Dilemma.} Representing mathematical expressions faces a fundamental trade-off. On one hand, generating linear LaTeX string-based representation is straightforward. However, it complicates the encoding of the complex layout of mathematical expressions. On the other hand, syntax tree-based representation, while intuitive, poses significant integration challenges with the now dominant, autoregressive decoding models.

Facing these challenges, in this paper we present three contributions to complex MER. Firstly, we introduce CMER-Bench, a benchmark with a clear difficulty stratification, designed to enable rigorous evaluation of model performance across Easy, Moderate, and Complex levels. Here Easy roughly corresponds to the difficulty of existing datasets. Secondly, we introduce two training datasets, CMER-3M and MER-17M, the former being the most complex and the latter the largest MER training dataset to our knowledge. Both are larger than existing MER training datasets to our knowledge. We have made it publicly available to support the development MER models for the community. 
Thirdly, we develop CMERNet, a encoder-decoder-based baseline with new tokenizer and representation. When trained on CMER-3M and evaluated on CMER-Bench, CMERNet achieves state-of-the-art (SOTA) performance across a majority of evaluation metrics, providing the community with a powerful model to comprehensively address the challenges of complex MER.

\section{Related Work}
Early deep learning-based methods treat MER as an image-to-sequence interpretation task, typically using an attention-based CNN encoder and an RNN decoder~\cite{ZHANG2017196,deng2017im2markup}. IM2LATEX \cite{deng2017im2markup} exemplifies this image-to-sequence approach and introduces the IM2LATEX-100K dataset to support MER model training. This architecture is later enhanced by Pix2tex \cite{Blecher2022latexocr}, which replaces the CNN with a Vision Transformer (ViT) \cite{dosovitskiy2021imageworth16x16words} and the RNN with a Transformer decoder to capture rich global dependencies. To address the issue of lacking training data, \citeauthor{wang_unimernet_2024} introduces the large-scale UniMER-1M dataset with 1M instances. They also develop the UniMERNet model that achieves significant performance gains when trained on the dataset. These methods are not good at describing the spatial layout of expressions. The linear nature of LaTeX string introduces ambiguity upon tokenization, making structural tokens like \verb|{|, \verb|}|, and \verb|\frac| challenging to interpret. 


Consequently, some research has shifted towards predicting more structured formats, such as syntax trees. TDv2 \cite{Wu_tdv2_2022} introduces a tree-structured decoder that autoregressively constructs a formula's syntax tree. It decouples each generation step into two parallel sub-tasks: to predict the symbol's identity and to determine its spatial relationship to the parent. Another tree-based method, TAMER \cite{zhu_tamer_2024}, enhances the decoder by jointly modeling the symbol sequence and its hierarchical structure through tree-aware attention. However, both methods are mainly suitable for expressions with relatively simple layouts and fewer tokens, such as handwritten mathematical expressions~\cite{Mouchère2013ICDAR,mouchere2016icfhr2016,xie2023icdar}.

\begin{figure}[t]
\centering 
    \includegraphics[width=0.48\textwidth]{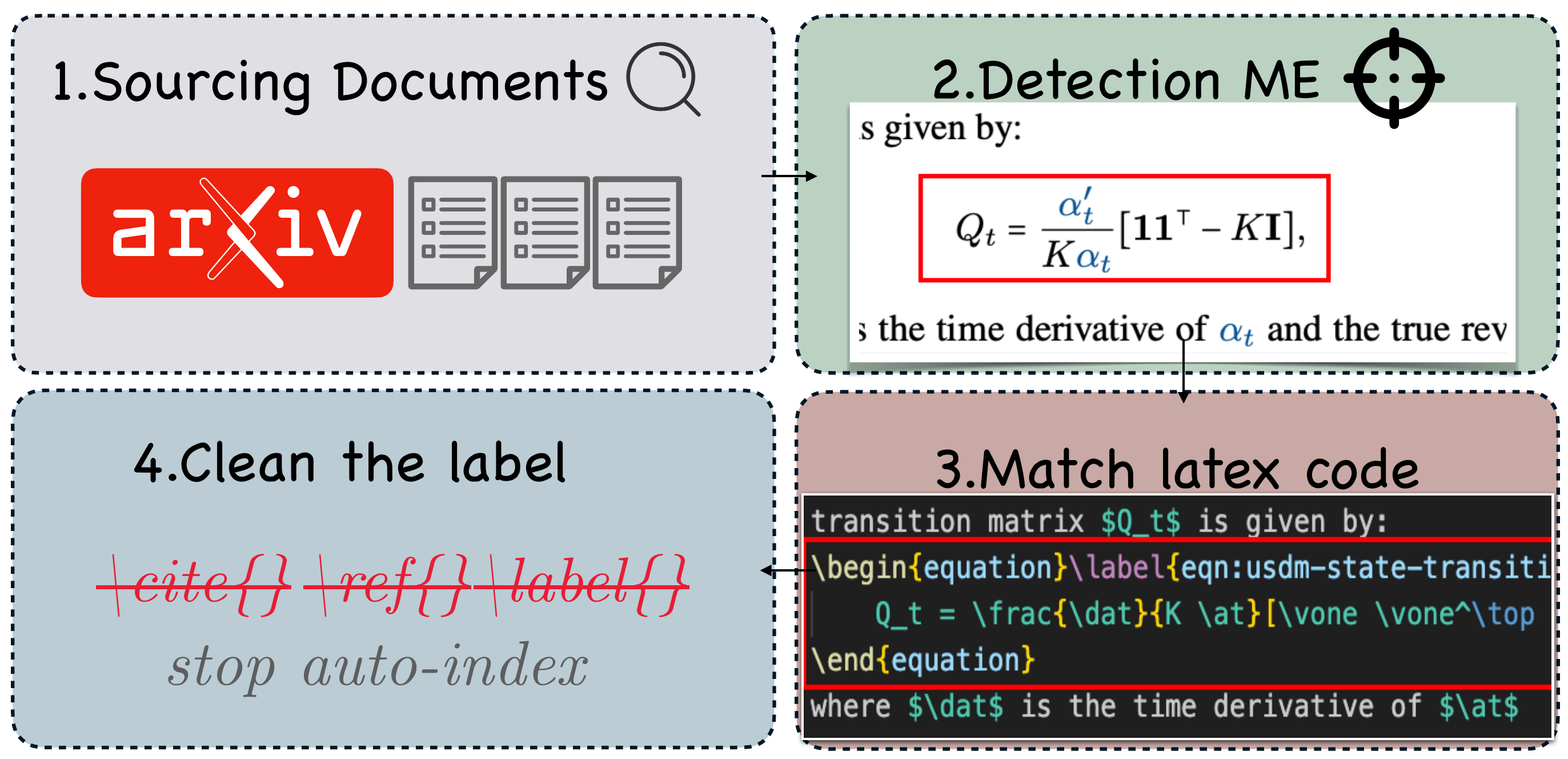}
\caption{The pipeline of constructing CMER-3M and MER-17M.}
\label{fig:5}
\end{figure}

\begin{table}[t]
\centering
{%
\small            
\setlength{\tabcolsep}{1pt}
\begin{tabular}{lccccc}
\toprule
Length   & IM2LATEX & Pix2tex & UniMER    & CMER-3M & MER-17M   \\
\midrule
$0-20$    & $<$0.1K & 3.6K   & \textbf{477.8K}   & 85.9K  & 85.9K   \\
$21-150$  & 43.4K   & 137.5K & 342.6K   & 620.2K & \textbf{5.5M}  \\
$151-300$ & 28.1K   & 70.6K  & 103.1K   & 858.1K & \textbf{7.8M}  \\
$301-450$ & 3.6K    & 16.8K  & 62.6K    & 855.6K & \textbf{3.0M}   \\
$>450$    & 0.1K    & 5.8K   & 75.6K    & 641.2K & \textbf{1.4M}  \\
\midrule
Total     & 75.3K   & 233.8K & 1.1M     & 3.1M   & \textbf{17.7M}\\
\bottomrule
\end{tabular}
}
\caption{Statistics of current MER datasets and ours, categorized by the length of LaTeX sequences. }
\label{tab:dataset_stats_final}
\end{table}

\begin{table}[t]
\centering
{%
\small            
\setlength{\tabcolsep}{1pt}
\begin{tabular}{lccccc}
\toprule
Lines & IM2LATEX & Pix2tex & UniMER    & CMER-3M & MER-17M   \\
\midrule
$1$     & 71.7K   & 221.8K & \textbf{964.7K}   & 505.7K & 504.3K  \\
$2$     & 0.3K    & 6.0K   & 45.6K    & 2.0M & \textbf{15.6M} \\
$3$     & 2.0K    & 3.6K   & 23.7K    & 407.4K & \textbf{1.3M}   \\
$4$     & 0.4K    & 1.6K   & 13.1K    & 52.5K & \textbf{143.3K}   \\
$5$     & 0.5K    & 0.8K     & 5.8K     & 25.2K  & \textbf{60.6K}  \\
$>5$    & 0.3K    & 1.1K   & 8.9K     & 21.9K & \textbf{43.5K}   \\
\bottomrule
\end{tabular}
}
\caption{Expression distribution based on the number of lines they occupied.}
\label{tab:line_distribution_final}
\end{table}

\section{Complex MER Datasets and Benchmark}

\subsection{CMER-3M and MER-17M Datasets}

Existing public datasets for MER suffer from clear limitations, including insufficient data scale, mainly composed of structurally simple and single-line expressions, etc, restricting their usage as comprehensive MER training corpora. To overcome the issues, we introduce MER-17M, a large-scale dataset for MER, and its subset CMER-3M that emphasizes both distribution balance and data complexity. MER-17M is constructed through a multi-stage pipeline that begins with sourcing over 1 million scientific documents from repositories like ArXiv \cite{arxiv}. We extract mathematical expressions using advanced parsing tools to isolate LaTeX code snippets, clean those visual-semantic misalignment tags like \verb|\cite|, which confuse the model. After that we render them into images. A rigorous filtering process ensures its quality: we discard malformed or overly simplistic expressions and apply diversity checks. As a result, we obtain over 17 million expression image-LaTeX pairs. We choose 2000 of them as the evaluation benchmark, and the collection of the rest is termed MER-17M.

Since the MER-17M distribution is not balanced, we further construct CMER-3M, a subset of MER-17M, which is more balanced across expression length. CMER-3M is featured in the following three key dimensions:  


\textbf{Unprecedented scale.} CMER-3M largely surpasses prior datasets in instance quantity, containing over 3 million samples—nearly 3 times larger than UniMER (1M), the current largest, and much larger than IM2LATEX (100K) or Pix2tex (234K), as quantified in Table~\ref{tab:dataset_stats_final}. This scale ensures many rare symbols and intricate patterns are included, building a strong foundation for robust model learning. 

\textbf{Balanced length distribution.} Unlike existing datasets biased towards short-length (in LaTeX token) expressions, CMER-3M is more uniform in length distribution. As illustrated in Table~\ref{tab:dataset_stats_final}, CMER-3M contains much more medium- and long-length expressions. Since length is the most important indicator describing the complexity of expressions, the existence of rich complex expressions makes complex MER studies no longer troubled by training data.

\textbf{High structural complexity.} As shown in Table \ref{tab:line_distribution_final}, CMER-3M has more than 2.5M multi-line expressions, accounting for 83\% of its samples. The ratio significantly exceeds existing datasets, especially in expressions with three or more lines. The statistics also indicate that MER-17M is rich in diverse and complex layouts, which is also a critical distinction to existing datasets.  


\begin{figure*}[!t]
\centering 
    \includegraphics[width=0.98\textwidth]{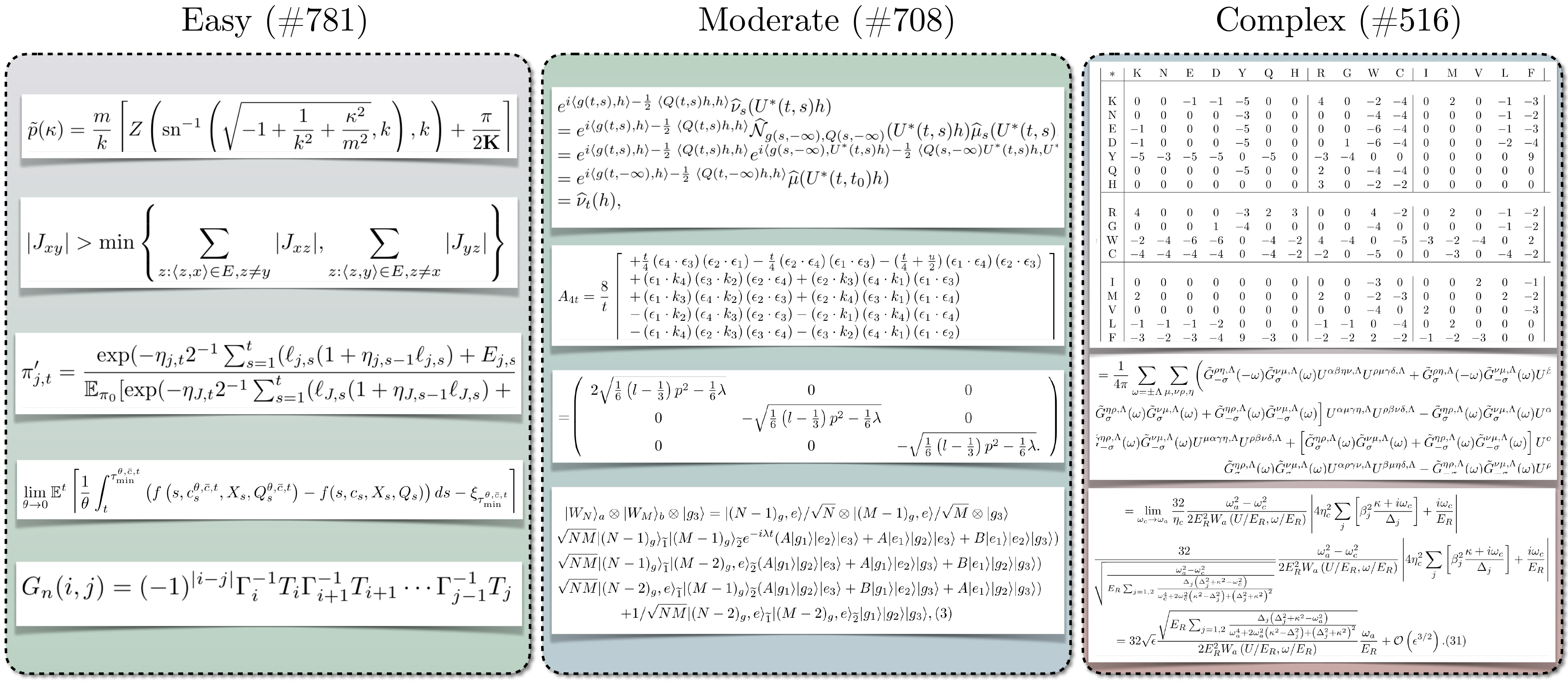}
\caption{Showcase of CMER-Bench examples according to their difficulty tiers.}
\label{fig:6}
\end{figure*}

\subsection{CMER-Bench Benchmark}

Complex MER is also hindered by the scarcity of comprehensive evaluation benchmarks. Existing test sets are heavily skewed towards simple, single or two-line, or synthetic expressions. These sets only partially evaluate the capability of existing models, where their ability in handling complex MER has not been fully assessed.


To address this issue, we introduce CMER-Bench, a benchmark meticulously curated to facilitate a comprehensive and fair assessment of MER models. While also derived from the multi-stage pipeline described in Sec. 3.1, the composition of CMER-Bench is determined through a rigorous hybrid process, combining automated difficulty scoring with human expert verification to ensure both the quality and representativeness of selected samples. Following this criterion, 2000 samples are selected. 

We divide CMER-Bench into three distinct subsets: Easy, Moderate, and Complex, indicating increasing recognition difficulties. Examples and numbers of each tier are shown in Figure \ref{fig:6}. This classification enables us to observe the performance of different MER models along the difficulty axis, supporting more comprehensive model assessment.


\section{CMERNet}

To develop a general yet powerful MER model, we propose specialized tokenizer, expression representation, and modeling methods as follows.

\subsection{Specialized Mathematical Tokenizer}

As shown by studies in scientific text domain  \cite{taylor2022galacticalargelanguagemodel}, using custom tokenizer to parse the syntax of specified data can obtain better results. Mathematical expressions have their own syntax. However, no specialized tokenizer for MER has been proposed to date. Using general-purpose tokenizers often likely fragment critical mathematical commands (e.g., splitting \verb|\sqrt| into \verb|\|, s, q, r, t), which disrupts semantic integrity and complicates model learning.


To address this, we develop a specialized Byte Pair Encoding (BPE) tokenizer. It parses the expression syntax following a two-step process. First, we curate a comprehensive vocabulary of common LaTeX environments and commands, such as \verb|\sqrt|, \verb|\frac|, and \verb|\begin{gather}|. These were explicitly designated as special tokens, ensuring that they are treated as atomic units and preventing fragmentation. Second, we trained the BPE tokenizer on our CMER-3M dataset. This process yields a highly efficient mathematical tokenizer that preserves structural semantics well.


\subsection{Structured Mathematical Language (SML)}
To overcome the inherent limitation of linear LaTeX string-based representation, we introduce a novel representation called SML that explicitly injects structural information into the expression representation such that benefits learning process. Our goal is to create a target sequence that both preserves the rich, two-dimensional grammar of mathematical expressions, and seamlessly integrates these with common autoregressive decoder architectures.
To achieve this, we have developed a two-step conversion scheme as follows:

\textbf{Parsing the LaTeX string into the syntax tree.} First, we parse the raw LaTeX string into a hierarchical syntax tree. This tree explicitly captures the nested relationships and the parent-child grammar between operators and operands, e.g., identifying the numerator and denominator within a fraction. This tree encodes the spatial structure of the expression.

\textbf{Serializing the tree into a structured token sequence by additional grammar tokens.} Then, we traverse this syntax tree to serialize it into a sequence of tokens. Note that this is not a flat representation like LaTeX string. Instead, the token sequence directly encodes the tree's structure using additional designated tokens to signify parent-child relationships, node types, and sibling order.

The SML representation offers two advantages. First, it resolves the representation bottleneck of LaTeX by providing the model with a structure-aware target that makes the correspondence between visual layout and token semantics explicit. Second, because the final output is still a linear sequence of tokens---albeit a structurally meaningful one---it aligns seamlessly with the dominant autoregressive decoder. This makes our method highly scalable and allows it to directly benefit from advances in current leading models, facilitating our model design.

\begin{figure}[t]
\centering 
    \includegraphics[width=0.48\textwidth]{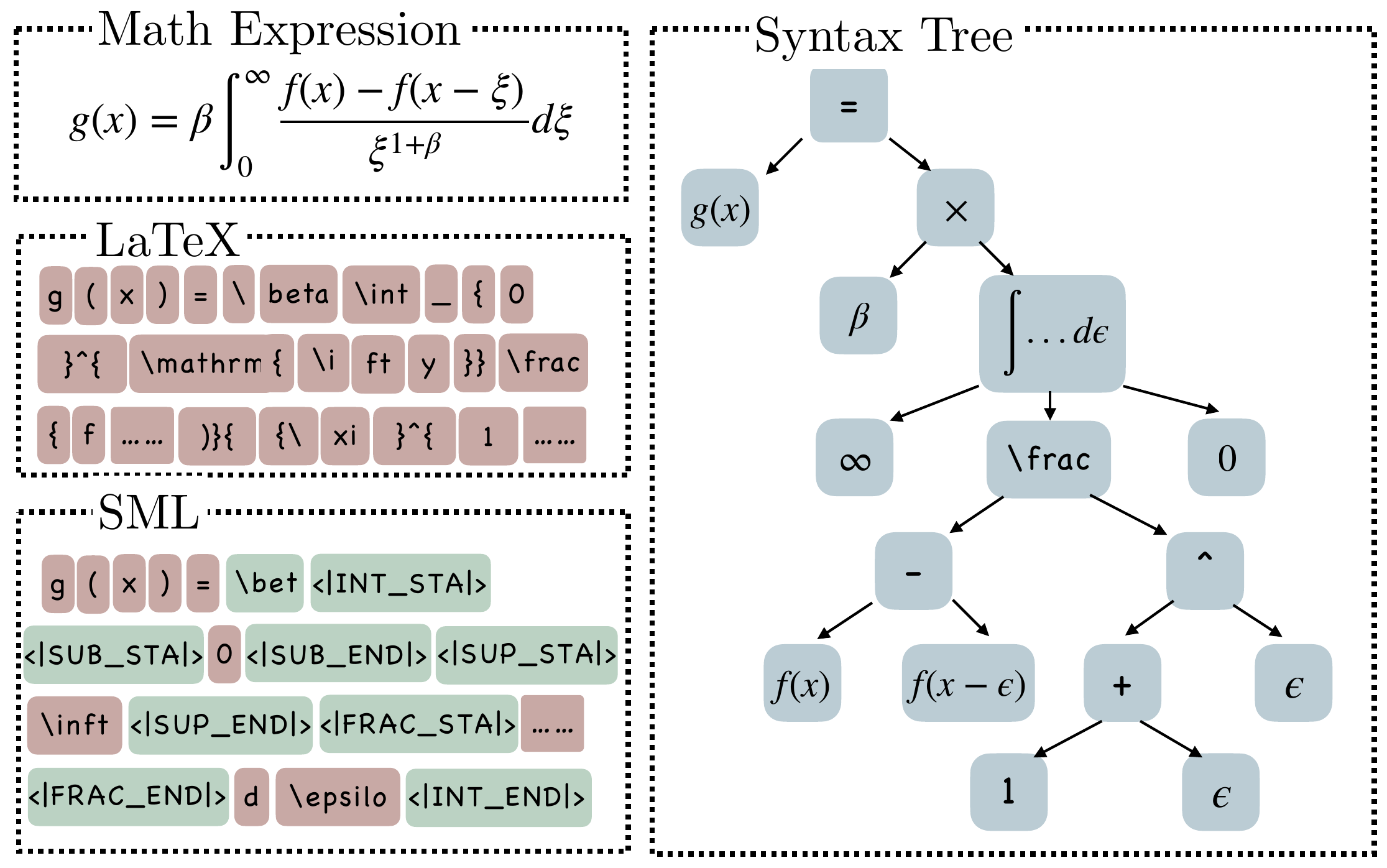}
\caption{An example of syntax tree-based expression representation. Note that the raw LaTeX string can be represented as SML following the syntax tree.}
\label{fig:4}
\end{figure}

\begin{figure*}[!t]
\centering 
    \includegraphics[width=0.98\textwidth]{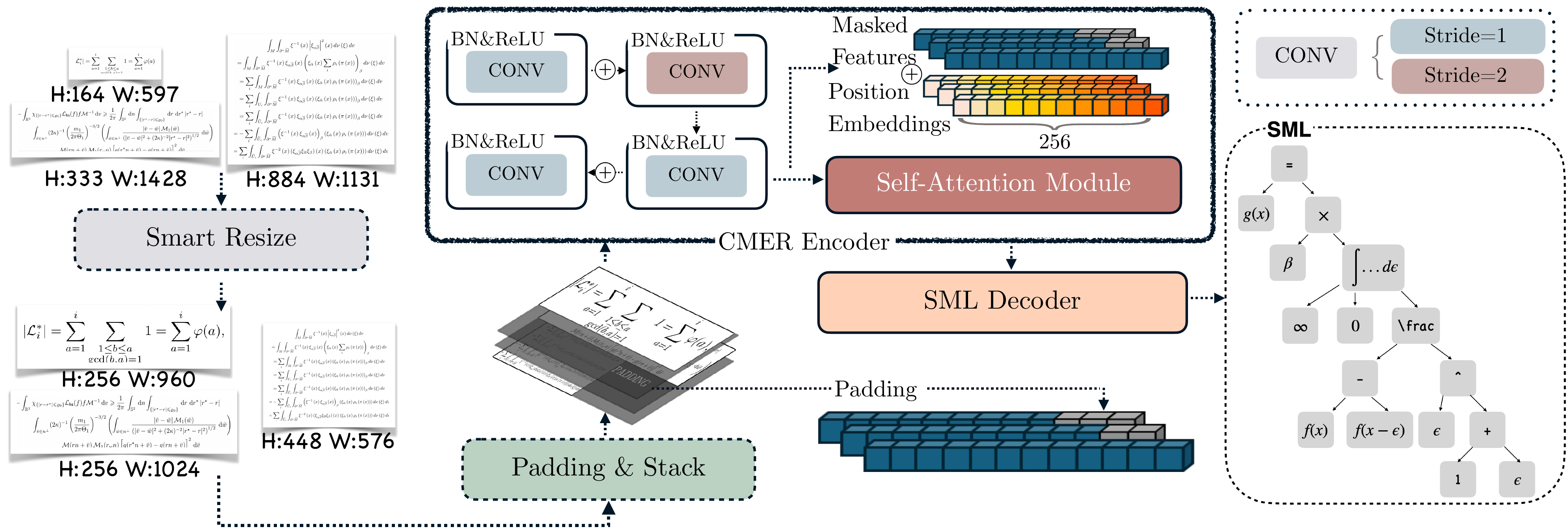}
\caption{The architecture of CMERNet. Our model employs a hybrid vision encoder that sequentially combines a CNN for fine-grained feature extraction with a Transformer backbone for global context modeling. This design is tailored to robustly process complex mathematical expressions with both intricate local details and long-range structural dependencies.}
\label{fig:2}
\end{figure*}

\subsection{Dynamic Resolution Pre-processing}

Mathematical expressions have rich fine-grained details and often non-standard, portrait-oriented aspect ratios. Thus, they are highly susceptible to distortion when using conventional fixed-sized image pre-processing, e.g., downsampling. To address this, we introduce CMER-Fit, a dynamic resolution adjustment module inspired by NaViT \cite{dehghani2023patchnpacknavit} and SigLip2 \cite{tschannen2025siglip2multilingualvisionlanguage}.

CMER-Fit's objective is to render each image using the highest resolution that adheres to a predefined token budget (e.g., 256 patches) required by the vision encoder. Given an original image of size $H_{0}\!\times\! W_{0}$ and a patch size $p\times p$, we seek the greatest scale factor $s^{\star}$ such that the resulting patch grid does not exceed $N_{\max}$ elements, $H^*$ and $W^*$ are the adjusted height and width, respectively.
\begin{align}
\Bigl\lceil \tfrac{s\,H_{0}}{p}\Bigr\rceil
\;\times\;
\Bigl\lceil \tfrac{s\,W_{0}}{p}\Bigr\rceil
\;\le\; N_{\max}\\
H^{\star} = p \Bigl\lceil \tfrac{s^{\star} H_{0}}{p}\Bigr\rceil,\quad
W^{\star} = p \Bigl\lceil \tfrac{s^{\star} W_{0}}{p}\Bigr\rceil
\end{align}


By using dynamic resolution resizing, we can explicit measure the distortion caused by this resizing, which we define as Maximum Distortion Ratio (MDR) as follows.
\begin{align}
    \mathrm{MDR}=1-\frac{H^{\star}W^{\star}}{H_{0}W_{0}}
\end{align}

Note that we only consider downsampling-based resizing. That is, if $H_{0}W_{0}\le p^{2}N_{\max}$, this means $s^{\star}\ge1$ and corresponds to upsampling. In this case, no resizing occurs and $\operatorname{MDR}=0$.

Because each side is rounded up to a multiple of $p$, at most $p-1$ surplus pixels are introduced along each axis. Hence
\begin{align}
0 \;\le\;
\mathrm{MDR}
= 1 - \frac{H^{\star}W^{\star}}{H_{0}W_{0}}
\;<\;
\frac{(p-1)(H_{0}+W_{0})}{H_{0}W_{0}}.
\end{align}

Setting $H_{0}=W_{0}\gg p$ gives the theoretical upper bound
\begin{align}
\mathrm{MDR}_{\max}<\frac{2(p-1)}{H_{0}}.
\end{align}

\begin{table}[htb]
\centering
\small

\begin{tabular}{ccc}
\toprule
Patch size $p$ & Typical input $H_{0}=W_{0}$ & $\mathrm{MDR}_{\max}$ \\
\midrule
$16$ & $1024$ & $<2.93\%$ \\
$32$ & $1024$ & $<6.05\%$ \\
\bottomrule
\end{tabular}
\caption{Maximum distortion ratio under different patch sizes with respect to $1024\times1024$ input.}
\label{tab:mdr}
\end{table}

Two practical examples shown in Table \ref{tab:mdr} illustrate the scale of this bound. It is observed that the distortion is negligible for high-resolution images.

CMER-Fit offers a parameter-free, theoretically sound mechanism for harmonizing token budget, memory overhead, and spatial fidelity. Its provably small worst-case distortion and negligible computational cost make it a robust choice for large-scale training and deployment under stringent hardware constraints.

\subsection{CMERNet: A Strong Baseline}
Building on the above techniques, we design CMERNet, a strong baseline model for complex MER. We develop specialized architecture to meet two key characteristics inherent in MER: capturing fine-grained visual details (e.g., dots, primes, small symbols) and understanding the complex, hierarchical 2D grammar of expressions. To this end, CMERNet is composed of three core components: a hybrid vision encoder, a feature-refining connector, and an autoregressive decoder. By incorporating SML, CMERNet could better understand mathematical expressions.

To effectively process both local details and global structure, our vision encoder adopts a CNN-Transformer hybrid design. Diverging from the traditional OCR model SVTRv2 \cite{du2025svtrv2}, and more complex interleaved architectures like CvT \citep{wu2021cvtintroducingconvolutionsvision} or task-specific models like DETR \citep{carion2020endtoendobjectdetectiontransformers}, our architecture explores the complementarity of well-established CNN and Transformer, leveraging their respective strengths in a two-step process:

\textbf{Step 1: CNN for fine-grained feature capture.} The vision encoder first employs a shallow CNN backbone composed of six stacked residual blocks. These convolutions are deliberately designed without downsampling in its early layers to preserve a high-resolution feature map. This is critical for capturing the precise location and appearance of fine-grained symbols like superscripts, subscripts, and diacritics, which are often lost with aggressive initial downsampling.

\textbf{Step 2: Transformer for global context modeling.} The CNN feature map is then passed to the Transformer part, where 12 standard Transformer layers are employed. This step is responsible for modeling the global context, allowing the model to understand the long-range dependencies and complex spatial relationships between all symbols in the expression.

Then, in the feature-refining connector, we adopt two MLP layers commonly used in MLLMs to align the image features extracted by vision encoder with the linguistic representations of mathematical formulas. Finally, an autoregressive decoder models these image features and decodes them into the SML representation. The decoder is built upon a cross-attention-based Transformer architecture.

\section{Experiments}

\subsection{Implementation Details}

\textbf{Training data.} The model is trained on our CMER-3M and a collection of existing datasets. To enhance the model's robustness and equip it with the capability to recognize handwritten mathematical expressions, in addition to CMER-3M, we augment our training data by incorporating samples from CROHME2013 \cite{Mouchère2013ICDAR}, CROHME2016 \cite{mouchere2016icfhr2016}, CROHME2019 \cite{mahdavi2019icdar} and CROHME2023 \cite{xie2023icdar}, along with other publicly available mathematical formula corpora such as HME-100K \cite{yuan2022syntax} and UniMER-1M \cite{wang_unimernet_2024}.


\textbf{Training Details.} We use the AdamW optimizer with default parameters ($\beta_1=0.9,\beta_2=0.999$). The training process spanned one full epoch over the dataset. We adopt a learning rate schedule with a linear warmup phase: the learning rate is increased from an initial value of $1 \times 10^{-5}$ to a peak of $ 1 \times 10^{-4}$ over the first 3,000 steps. Following the warmup, the learning rate is decayed to $1 \times 10^{-9}$ using a cosine annealing schedule for the remainder of the training. All experiments are conducted on a single server equipped with eight NVIDIA RTX 4090 GPUs. We employ Distributed Data Parallel (DDP) to accelerate training, setting a per-GPU batch size of 8, which results in a global batch size of 64. The whole training takes around 3 days. We train CMERNet from scratch using PyTorch. The whole model, comprising approximately 125 million parameters, is trained end-to-end.


\begin{table*}[t]
\newcolumntype{g}{>{\columncolor[HTML]{EFEFEF}}c} 
\footnotesize 
\centering

\setlength{\tabcolsep}{3.5pt} 
\begin{tabular}{l|ccc|cccccc}
\toprule
& \multicolumn{3}{c|}{\cellcolor[HTML]{ECF4FF}\textbf{Difficulty Level}} & \multicolumn{6}{c}{\cellcolor[HTML]{FFFFC7}\textbf{Performance Metrics}} \\
\cmidrule(lr){2-4} \cmidrule(lr){5-10}
\textbf{Model} & \textbf{Easy} & \textbf{Moderate} & \textbf{Complex} & \textbf{ROUGE-1}$\uparrow$ & \textbf{ROUGE-2}$\uparrow$ & \textbf{ROUGE-L}$\uparrow$ & \textbf{BLEU}$\uparrow$ & \textbf{Avg Edit Dist.} $\downarrow$ & \textbf{CDM} $\uparrow$ \\
\midrule

\multirow{3}{*}{UniMERNet \shortcite{wang_unimernet_2024}} 
& \cellcolor[HTML]{ECF4FF}\checkmark & & & 0.7464 & 0.6396 & 0.7388 & 0.4758 & 94.6978 & 0.938 \\
& & \cellcolor[HTML]{ECF4FF}\checkmark & & 0.68 & 0.5449 & 0.6703 & 0.4575 & 133.2556 & 0.866 \\
& & & \cellcolor[HTML]{ECF4FF}\checkmark & 0.7605 & \textbf{0.6126} & 0.7502 & 0.4093 & 483.34 & 0.68 \\
\midrule

\multirow{3}{*}{Gemini-2.5-pro \shortcite{gemini2025pro}} 
& \cellcolor[HTML]{ECF4FF}\checkmark & & & 0.7402 & 0.6353 & 0.738 & 0.455 & 97.5134 & 0.862 \\
& & \cellcolor[HTML]{ECF4FF}\checkmark & & 0.5585 & 0.4029 & 0.5553 & 0.266 & 192.4393 & 0.862 \\
& & & \cellcolor[HTML]{ECF4FF}\checkmark & 0.6589 & 0.5424 & 0.6528 & 0.3089 & 1282.1165 & \textbf{0.856} \\
\midrule

\multirow{3}{*}{ChatGPT-4o \shortcite{gpt4o_system_card}} 
& \cellcolor[HTML]{ECF4FF}\checkmark & & & 0.7166 & 0.5972 & 0.7135 & 0.4517 & 145.306 & 0.888 \\
& & \cellcolor[HTML]{ECF4FF}\checkmark & & 0.5418 & 0.3792 & 0.5372 & 0.185 & 200.6257 & 0.83 \\
& & & \cellcolor[HTML]{ECF4FF}\checkmark & 0.5977 & 0.486 & 0.5904 & 0.2082 & 1905.0465 & 0.635 \\
\midrule

\multirow{3}{*}{Claude-Sonnet-4 \shortcite{anthropic_claude_sonnet4_docs}} 
& \cellcolor[HTML]{ECF4FF}\checkmark & & & 0.7126 & 0.5996 & 0.7099 & 0.4341 & 131.3675 & 0.898 \\
& & \cellcolor[HTML]{ECF4FF}\checkmark & & 0.4905 & 0.3427 & 0.4864 & 0.1589 & 206.0297 & 0.747 \\
& & & \cellcolor[HTML]{ECF4FF}\checkmark & 0.6471 & 0.5356 & 0.6426 & 0.2623 & 1537.9671 & 0.671 \\
\midrule

\multirow{3}{*}{Qwen-VL-72b \shortcite{qwen2.5-VL}} 
& \cellcolor[HTML]{ECF4FF}\checkmark & & & 0.7048 & 0.5979 & 0.7023 & 0.4477 & 133.525 & 0.796 \\
& & \cellcolor[HTML]{ECF4FF}\checkmark & & 0.5371 & 0.3739 & 0.5279 & 0.2433 & 200.2345 & 0.742 \\
& & & \cellcolor[HTML]{ECF4FF}\checkmark & 0.5932 & 0.4935 & 0.5872 & 0.1915 & 2107.3029 & 0.792 \\
\midrule

\multirow{3}{*}{MinerUv2 \shortcite{wang2024mineruopensourcesolutionprecise}} 
& \cellcolor[HTML]{ECF4FF}\checkmark & & & 0.6238 & 0.4828 & 0.6218 & 0.333 & 144.1306 & 0.936 \\
& & \cellcolor[HTML]{ECF4FF}\checkmark & & 0.5716 & 0.4106 & 0.5676 & 0.3218 & 187.2429 & 0.866 \\
& & & \cellcolor[HTML]{ECF4FF}\checkmark & 0.6938 & 0.525 & 0.6863 & 0.3218 & 657.1495 & 0.63 \\
\midrule

\multirow{3}{*}{Dolphin \shortcite{feng2025dolphin}} 
& \cellcolor[HTML]{ECF4FF}\checkmark & & & 0.5911 & 0.4819 & 0.5832 & 0.3806 & 146.0141 & 0.901 \\
& & \cellcolor[HTML]{ECF4FF}\checkmark & & 0.6911 & 0.577 & 0.6689 & 0.4223 & 452.0213 & \textbf{0.886} \\
& & & \cellcolor[HTML]{ECF4FF}\checkmark & 0.6791 & 0.5488 & 0.6706 & 0.3805 & 776.9899 & 0.716 \\
\midrule

\multirow{3}{*}{CMERNet (Ours)} 
& \cellcolor[HTML]{ECF4FF}\checkmark & & & \textbf{0.8626} & \textbf{0.6613} & \textbf{0.8611} & \textbf{0.7653} & \textbf{51.61} & \textbf{0.966} \\
& & \cellcolor[HTML]{ECF4FF}\checkmark & & \textbf{0.8266} & \textbf{0.6072} & \textbf{0.8181} & \textbf{0.7215} & \textbf{88.09} & 0.845 \\
& & & \cellcolor[HTML]{ECF4FF}\checkmark & \textbf{0.7750} & 0.5764 & \textbf{0.7453} & \textbf{0.5568} & \textbf{474.18} & 0.690 \\

\bottomrule
\end{tabular}
\caption{Performance comparison of various models on CMER-Bench covering three difficulty levels. Our CMERNet demonstrates superior performance, especially on complex expressions. Best results in each column and difficulty level are in \textbf{bold}. }
\label{tab:main_results}
\end{table*}

\subsection{Evaluation Metrics}

\textbf{BLEU (Bilingual Evaluation Understudy).} 
BLEU \cite{papineni-etal-2002-bleu} computes sequence similarity using n-gram precision between a prediction and reference, supplemented by a brevity penalty to penalize short hypotheses. Although effective at capturing local correctness, its high sensitivity to the exact token sequence makes it less robust to synonymous LaTeX representation.

\textbf{ROUGE (Recall-Oriented Understudy for Gisting Evaluation).} 
We use ROUGE \cite{lin-2004-rouge}, reporting ROUGE-1 (unigram recall), ROUGE-2 (bigram recall), and ROUGE-L (longest common subsequence). ROUGE-L specifically rewards in-order, non-contiguous matches. As a string-level metric, it shares BLEU's susceptibility to the one-to-many mapping problem.

\textbf{Levenshtein Distance.}
Also known as edit distance \cite{haldar2011levenshteindistancetechniquedictionary}, this metric counts the minimum token-level insertions, deletions, and substitutions between prediction and reference. While intuitive (lower is better), its strict token-by-token comparison fails to recognize the structural or visual equivalence of synonymous LaTeX commands.

\textbf{CDM (Canonicalization-based Distance Measure).}
Proposed by \cite{wang2025imagetexttransformingformula}, CDM is a recent metric designed specifically to address one of many problems in MER. It first adds a color label in both predictions and labels, aiming to get the bounding box of each token in expressions. Then CDM apply matching algorithm to calculate f1 score and recall score. CDM can complement metrics such as BLEU to some extent.

BLEU and ROUGE are susceptible to length bias due to their n-gram match mechanism, which can obscure performance discrepancies, especially on complex expressions where similar scores may not reflect equal quality. We find that CDM, which evaluates structural element-wise similarity, can mitigate this issue. Therefore, we employ all these metrics, believing that this multi-faceted approach, balancing token overlap (BLEU/ROUGE) with structural fidelity, more accurately reflects true performance.

\subsection{Result Analysis}

We evaluate CMERNet against the baselines, including general-purpose MLLMs such as GPT-4o and Gemini-2.5-pro, state-of-the-art document parsing systems such as MinerU and Dolphin, and the dedicated MER method UniMERNet. This selection ensures a comprehensive comparison across different architectures and training objectives.

The results are listed in Table \ref{tab:main_results}. The baselines perform poorly on the complex tier. All compared methods, especially general-purpose MLLMs like ChatGPT-4o and Gemini, suffer from a drastic performance drop compared to our CMERNet. The BLEU scores for Gemini-2.5-pro, ChatGPT-4o, and QWen-VL-72b are only 0.3089, 0.2082, and 0.1915 on the complex tier, respectively. This confirms their inadequacy in handling intricate, multi-line structures. Furthermore, the expert model UniMERNet and document parsing models (MinerUv2 and Dolphin) get slightly better results, mainly because they are dedicatedly trained by expression data. Nevertheless, CMERNet, by training using a large number of mathematic expressions, as well as appropriately handling the tokenizer and representation issue, excelling the second best method in Table \ref{tab:main_results} nearly 15\%-29\% in the BLEU score under different difficulty tiers. Note that the CDM metric does not exhibit so large gaps, this is because CDM mainly measures the bounding box-level matching, which hardly perceive subtle mistakes. Nevertheless, the results clearly validate the effectiveness of our CMERNet.




\section{Conclusion}

In this work, our aim is to address the challenge of complex MER, a task that has been largely overlooked previously. Observing that complex MER still lacks comprehensive evaluation benchmark and large-scale training dataset, we first introduce CMER-Bench, a benchmark composed of easy, moderate and complex expressions, which enables the detailed assessment of the MER ability of different models, and for the first time, covers the evaluations of complex expressions. Then, we construct CMER-3M and MER-17M, both excel existing datasets in instance quantity and diversity. We also prove that the dataset is more suitable to train robust MER models. Furthermore, we have investigated the expression tokenizer and representation, and develop a specialized MER model termed CMERNet. Experimental results demonstrate that CMERNet, despite only 125M parameters, has shown strong performance compared to general-purpose MLLMs, document parsing models and specialized MER model. The results basically validate the importance of introducing CMER-Bench, CMER-3M and MER-17M, and the effectiveness of our CMERNet. In future, we plan to conduct more experiments to further verify CMERNet, as well as exploring using the constructed datasets to strengthen the MER capability of popular MLLMs.


\section*{Acknowledgments}
This work was supported by National Natural Science Foundation of China (No. 32341012).

\bibliography{aaai2026}

@article{zhu_tamer_2024,
  author    = {Zhu, Jianhua and Zhao, Wenqi and Li, Yu and Hu, Xingjian and Gao, Liangcai},
  title     = {{TAMER}: Tree-Aware Transformer for Handwritten Mathematical Expression Recognition},
  journal   = {arXiv preprint arXiv:2408.08578},
  year      = {2024}
}

@article{wang_unimernet_2024,
  author    = {Wang, Bin and Gu, Zhuangcheng and Liang, Guang and Xu, Chao and Zhang, Bo and Shi, Botian and He, Conghui},
  title     = {{UniMERNet}: A Universal Network for Real-World Mathematical Expression Recognition},
  journal   = {arXiv preprint arXiv:2404.15254},
  year      = {2024}
}

@inproceedings{Wu_tdv2_2022,
  author    = {Wu, Changjie and Du, Jun and Li, Yunqing and Zhang, Jianshu and Yang, Chen and Ren, Bo and Hu, Yiqing},
  title     = {{TDv2}: A Novel Tree-Structured Decoder for Offline Mathematical Expression Recognition},
  booktitle = {{AAAI}},
  volume    = {36},
  number    = {3},
  year      = {2022}
}

@inproceedings{li2025pacm,
  title={PACM: Position-Aware Cross-Modality Decoder for Handwritten Mathematical Expression Recognition},
  author={Li, Zeng and Wei, Jin and Shen, Zhijie and Ma, Can and Wu, Yaqiang and Zhou, Yu},
  booktitle={ICDAR},
  pages={96--114},
  year={2025},
}

@inproceedings{deng2017im2markup,
  author    = {Deng, Yuntian and Kanervisto, Anssi and Ling, Jeffrey and Rush, Alexander M.},
  title     = {Image-to-Markup Generation with Coarse-to-Fine Attention},
  booktitle = {{ICML}},
  pages     = {980--989},
  year      = {2017}
}

@misc{Blecher2022latexocr,
  author       = {Blecher, Lukas},
  title        = {pix2tex: {LaTeX}-{OCR} -- Using a {ViT} to Convert Images of Equations into {LaTeX} Code},
  year         = {2021},
  howpublished = {https://github.com/lukas-blecher/LaTeX-OCR}
}

@article{dehghani2023patchnpacknavit,
  author    = {Dehghani, Mostafa and Mustafa, Basil and Djolonga, Josip and others},
  title     = {Patch n' Pack: {NaViT}, a Vision Transformer for Any Aspect Ratio and Resolution},
  journal   = {arXiv preprint arXiv:2307.06304},
  year      = {2023}
}

@article{tschannen2025siglip2multilingualvisionlanguage,
  author    = {Tschannen, Michael and Gritsenko, Alexey and Wang, Xiao and others},
  title     = {{SigLIP} 2: Multilingual Vision-Language Encoders with Improved Semantic Understanding, Localization, and Dense Features},
  journal   = {arXiv preprint arXiv:2502.14786},
  year      = {2025}
}

@article{wu2021cvtintroducingconvolutionsvision,
  author    = {Wu, Haiping and Xiao, Bin and Codella, Noel and others},
  title     = {{CvT}: Introducing Convolutions to Vision Transformers},
  journal   = {arXiv preprint arXiv:2103.15808},
  year      = {2021}
}

@article{carion2020endtoendobjectdetectiontransformers,
  author    = {Carion, Nicolas and Massa, Francisco and Synnaeve, Gabriel and others},
  title     = {End-to-End Object Detection with Transformers},
  journal   = {arXiv preprint arXiv:2005.12872},
  year      = {2020}
}

@InProceedings{du2025svtrv2,
    author    = {Du, Yongkun and Chen, Zhineng and Xie, Hongtao and Jia, Caiyan and Jiang, Yu-Gang},
    title     = {SVTRv2: CTC Beats Encoder-Decoder Models in Scene Text Recognition},
    booktitle = {ICCV},
    year      = {2025},
    pages     = {20147-20156}
}

@article{du2025instruction,
  title={Instruction-guided scene text recognition},
  author={Du, Yongkun and Chen, Zhineng and Su, Yuchen and Jia, Caiyan and Jiang, Yu-Gang},
  journal={IEEE Trans. Pattern Anal. Mach. Intell.}, 
  year={2025},
  volume={47},
  number={4},
  pages={2723--2738},
}

@inproceedings{du2022svtr,
  title     = {{SVTR}: Scene Text Recognition with a Single Visual Model},
  author    = {Du, Y. and Chen, Z. and Jia, C. and Yin, X. and Zheng, T. and Li, C. and Du, Y. and Jiang, Y-G.},
  booktitle = {IJCAI},
  pages     = {884--890},
  year      = {2022},
}

@article{zheng2024cdistnet,
  title={Cdistnet: Perceiving multi-domain character distance for robust text recognition},
  author={Zheng, Tianlun and Chen, Zhineng and Fang, Shancheng and Xie, Hongtao and Jiang, Yu-Gang},
  journal={IJCV},
  volume={132},
  number={2},
  pages={300--318},
  year={2024},
}

@article{du2025context,
  title={Context perception parallel decoder for scene text recognition},
  author={Du, Yongkun and Chen, Zhineng and Jia, Caiyan and Yin, Xiaoting and Li, Chenxia and Du, Yuning and Jiang, Yu-Gang},
  journal={IEEE Trans. Pattern Anal. Mach. Intell.}, 
  year={2025},
  volume={47},
  number={6},
  pages={4668--4683},
}

@inproceedings{papineni-etal-2002-bleu,
  author    = {Papineni, Kishore and Roukos, Salim and Ward, Todd and Zhu, Wei{-}Jing},
  title     = {{BLEU}: A Method for Automatic Evaluation of Machine Translation},
  booktitle = {{ACL}},
  pages     = {311--318},
  year      = {2002}
}

@inproceedings{lin-2004-rouge,
  author    = {Lin, Chin{-}Yew},
  title     = {{ROUGE}: A Package for Automatic Evaluation of Summaries},
  booktitle = {{ACL} Workshop on Text Summarization},
  pages     = {74--81},
  year      = {2004}
}

@article{haldar2011levenshteindistancetechniquedictionary,
  author    = {Haldar, Rishin and Mukhopadhyay, Debajyoti},
  title     = {Levenshtein Distance Technique in Dictionary Lookup Methods: An Improved Approach},
  journal   = {arXiv preprint arXiv:1101.1232},
  year      = {2011}
}

@inproceedings{wang2025imagetexttransformingformula,
  author    = {Wang, Bin and Wu, Fan and Ouyang, Linke and Gu, Zhuangcheng and Zhang, Rui and Xia, Renqiu and Shi, Botian and Zhang, Bo and He, Conghui},
  title     = {Image Over Text: Transforming Formula Recognition Evaluation with Character Detection Matching},
  booktitle = {{CVPR}},
  pages     = {19681--19690},
  year      = {2025}
}

@article{wang2024mineruopensourcesolutionprecise,
  author    = {Wang, Bin and Xu, Chao and Zhao, Xiaomeng and others},
  title     = {MinerU: An Open-Source Solution for Precise Document Content Extraction},
  journal   = {arXiv preprint arXiv:2409.18839},
  year      = {2024}
}

@misc{qwen2.5-VL,
  author       = {Qwen Team},
  title        = {Qwen2.5-{VL}},
  year         = {2025},
  howpublished = {https://qwenlm.github.io/blog/qwen2.5-vl/}
}

@article{feng2025dolphin,
  author    = {Feng, Hao and Wei, Shu and Fei, Xiang and others},
  title     = {Dolphin: Document Image Parsing via Heterogeneous Anchor Prompting},
  journal   = {arXiv preprint arXiv:2505.14059},
  year      = {2025}
}

@article{gemini2025pro,
  author    = {{Gemini Team}},
  title     = {Gemini 2.5: Pushing the Frontier with Advanced Reasoning, Multimodality, Long Context, and Next Generation Agentic Capabilities},
  journal   = {arXiv preprint arXiv:2507.06261},
  year      = {2025}
}

@techreport{gpt4o_system_card,
  author       = {Hurst, Aaron and others},
  title        = {GPT-4o System Card},
  institution  = {OpenAI},
  year         = {2024}
}

@misc{anthropic_claude_sonnet4_docs,
  author       = {{Anthropic}},
  title        = {Models Overview: Claude Sonnet 4},
  year         = {2025},
  howpublished = {Anthropic Documentation}
}

@article{dosovitskiy2021imageworth16x16words,
  author    = {Dosovitskiy, Alexey and Beyer, Lucas and Kolesnikov, Alexander and others},
  title     = {An Image Is Worth 16x16 Words: Transformers for Image Recognition at Scale},
  journal   = {arXiv preprint arXiv:2010.11929},
  year      = {2021}
}

@misc{arxiv,
  author       = {{Cornell University}},
  title        = {arXiv: e-Print Archive},
  year         = {2025},
  howpublished = {https://arxiv.org}
}

@article{taylor2022galacticalargelanguagemodel,
  author    = {Taylor, Ross and Kardas, Marcin and Cucurull, Guillem and others},
  title     = {Galactica: A Large Language Model for Science},
  journal   = {arXiv preprint arXiv:2211.09085},
  year      = {2022}
}

@inproceedings{mouchere2016icfhr2016,
  author    = {Mouch{\`e}re, Harold and Viard{-}Gaudin, Christian and Zanibbi, Richard and Garain, Utpal},
  title     = {ICFHR2016 {CROHME}: Competition on Recognition of Online Handwritten Mathematical Expressions},
  booktitle = {{ICFHR}},
  pages     = {607--612},
  year      = {2016}
}

@inproceedings{Mouchère2013ICDAR,
  author    = {Mouch{\`e}re, Harold and Viard{-}Gaudin, Christian and Zanibbi, Richard and Garain, Utpal and Kim, Dae and Kim, Jin},
  title     = {ICDAR 2013 {CROHME}: Third International Competition on Recognition of Online Handwritten Mathematical Expressions},
  booktitle = {{ICDAR}},
  year      = {2013}
}

@inproceedings{xie2023icdar,
  author    = {Xie, Yejing and Mouch{\`e}re, Harold and Liwicki, Foteini Simistira and others},
  title     = {ICDAR 2023 {CROHME}: Competition on Recognition of Handwritten Mathematical Expressions},
  booktitle = {{ICDAR}},
  pages     = {553--565},
  year      = {2023}
}

@inproceedings{mahdavi2019icdar,
  author    = {Mahdavi, Mahshad and Zanibbi, Richard and Mouch{\`e}re, Harold and Viard{-}Gaudin, Christian and Garain, Utpal},
  title     = {ICDAR 2019 {CROHME+TFD}: Competition on Recognition of Handwritten Mathematical Expressions and Typeset Formula Detection},
  booktitle = {{ICDAR}},
  pages     = {1533--1538},
  year      = {2019}
}

@article{yuan2022syntax,
  author    = {Yuan, Ye and Liu, Xiao and Dikubab, Wondimu and others},
  title     = {Syntax-Aware Network for Handwritten Mathematical Expression Recognition},
  journal   = {arXiv preprint arXiv:2203.01601},
  year      = {2022}
}

@article{ZHANG2017196,
title = {Watch, attend and parse: An end-to-end neural network based approach to handwritten mathematical expression recognition},
journal = {Pattern Recognition},
pages = {196-206},
year = {2017},
author = {Jianshu Zhang and Jun Du and Shiliang Zhang and Dan Liu and Yulong Hu and Jinshui Hu and Si Wei and Lirong Dai}
}


\end{document}